\documentclass[letterpaper, 10 pt, conference]{ieeeconf}  % Comment this line out if you need a4paper 

\IEEEoverridecommandlockouts                              % This command is only needed if  
\usepackage{graphics} % for pdf, bitmapped graphics files
\usepackage{epsfig} % for postscript graphics files
\usepackage{mathptmx} % assumes new font selection scheme installed
\usepackage{times} % assumes new font selection scheme installed
\usepackage{amsmath} % assumes amsmath package installed
\usepackage{amssymb}  % assumes amsmath package installed
\usepackage[utf8]{inputenc} % allow utf-8 input
\usepackage[T1]{fontenc}    % use 8-bit T1 fonts
\usepackage{hyperref}       % hyperlinks
\usepackage{url}            % simple URL typesetting
\usepackage{booktabs}       % professional-quality tables
\usepackage{amsfonts}       % blackboard math symbols
\usepackage{nicefrac}       % compact symbols for 1/2, etc.
\usepackage{microtype}      % microtypography
\usepackage{xcolor}         % colors
\usepackage{cleveref}
\usepackage{graphicx}
\usepackage{float}
\usepackage{multirow}
\usepackage{diagbox}

\usepackage{algorithm}
\usepackage{algpseudocode}
\usepackage{arydshln}
\usepackage{marvosym}

\usepackage{multirow}
\usepackage{listings}
\usepackage{float}
\usepackage{siunitx} % \si{\degree}
\definecolor{dkgreen}{rgb}{0,0.6,0}
\definecolor{gray}{rgb}{0.5,0.5,0.5}
\definecolor{mauve}{rgb}{0.58,0,0.82}
\title{\LARGE \bf
RGBManip: Monocular Image-based Robotic Manipulation \\ through Active Object Pose Estimation % during Robot Arm Movement
}
\author{Boshi An$^{*}$, Yiran Geng$^{*}$, Kai Chen$^{*}$, Xiaoqi Li, Qi Dou, Hao Dong%~\textsuperscript{\Letter}
\thanks{Boshi An, Yiran Geng, Xiaoqi Li and Hao Dong are with Hyperplane Lab, School of CS, Peking University and National Key Laboratory for Multimedia Information Processing. 
Xiaoqi Li is also with Beijing Academy of Artificial Intelligence (BAAI).
Kai Chen and Qi Dou are with Department of Computer Science and Engineering, The Chinese University of Hong Kong.}
\thanks{* The first three authors contributed equally. }
\thanks{%\textsuperscript{\Letter} 
Corresponding to hao.dong@pku.edu.cn}
}

\begin{document}

\maketitle
\begin{abstract}
%Visual-based  
Robotic manipulation requires accurate perception of the environment, which poses a significant challenge due to its inherent complexity and constantly changing nature.
In this context, RGB image and point-cloud observations are two commonly used modalities
%~\cite{dai2022graspnerf,ichnowski2021dex,geng2022end,geng2023partmanip}
in visual-based robotic manipulation, but 
each of these modalities have their own limitations.
% Point-cloud observations suffer from sparsity in sampling and is unable to observe specular and transparent objects, while RGB images lack 3D information.
Commercial point-cloud observations often suffer from issues like sparse sampling and noisy output due to the limits of the emission-reception imaging principle.
%~\cite{xu2019survey}These challenges make it difficult to capture fine geometry details and observe specular or transparent objects. 
On the other hand, RGB images, while rich in texture information, lack essential depth and 3D information crucial for robotic manipulation. 
%To obtain such 3D information, the robot needs a way to capture the environment from different views.
%
%
To mitigate these challenges, we propose an image-only robotic manipulation framework that leverages an eye-on-hand monocular camera installed on the robot's parallel gripper.
% The eye-on-hand setting allow the robot to actively observe objects from different views for estimating its 6D pose while the manipulation process is on going.
%By adopting the eye-on-hand configuration, 
By moving with the robot gripper, this camera gains the ability to actively perceive the object from multiple perspectives during the manipulation process. This enables the estimation of 6D object poses, which can be utilized for manipulation.
% To manage the coordination between the manipulation policy and active perception, we employ a reinforcement learning policy to balance between the accuracy of 3D pose estimation and the time consumed during the manipulation process.
While, obtaining images from more and diverse viewpoints typically improves pose estimation, it also increases the manipulation time.
To address this trade-off, we employ a reinforcement learning policy to synchronize the manipulation strategy with active perception, achieving a balance between 6D pose accuracy and manipulation efficiency.
Our experimental results in both simulated and real-world environments showcase the state-of-the-art effectiveness of our approach. %, which, to the best of our knowledge, is the first to achieve robust real-world robotic manipulation through active pose estimation. 
We believe that our method will inspire further research on real-world-oriented robotic manipulation.
See \href{https://rgbmanip.github.io/}{https://rgbmanip.github.io/} for more details.
\end{abstract}

\section{Introduction}

% Robotic manipulation is a realm that holds tremendous potential for improving human lives. However, achieving robust and reliable robotic manipulation in our every-day life remains a significant challenge, due in part to the complexity and variability of the environment. One key issue is that the effectiveness of manipulation heavily depends on the ability of robots to perceive and understand the environment accurately.
Robotic manipulation is a field with immense potential to enhance human life. Nevertheless, realizing robust and dependable robotic manipulation in our daily lives continues to pose a substantial challenge, primarily due to the intricacies of our surroundings and the complexities in information acquisition. A critical factor in addressing this challenge lies in the precise perception and understanding of the environment by robots.

In this case, visual perception becomes a pivotal role in robotic manipulation, %as it enables robots to identify and locate objects, estimate their pose, and plan and execute manipulation tasks. 
as it facilitates object identification, localization, pose estimation, and task planning and execution.
In the quest for improved perception capabilities, researchers and engineers have commonly used RGB cameras and depth cameras as primary sources of sensory data~\cite{mo2021where2act,lv2022sagci,eisner2022flowbot3d,geng2022end}. 
% While both approaches have their strengths,
However, despite their respective strengths,
% they also have significant limitations that can impede their effectiveness
they both come with inherent drawbacks that may limit their applicability in complex or nuanced environments. Point-cloud data obtained from depth cameras, are often sparse and may fail to capture small or intricate features of objects, particularly at greater distances~\cite{kadambi20143d}. 
Additionally, although some industrial-grade depth cameras offer higher resolutions and extended capturing distances, these improvements often come at a significant financial cost, posing challenges for academic research and large-scale deployments~~\cite{horaud2016overview}. Moreover, depth cameras are prone to optical interference from other light sources and struggle with accurately imaging transparent and specular objects, such as glass and metal~\cite{kadambi20143d,xu2019survey,dai2022graspnerf}.
% RGB cameras, on the other hand, provides high-resolution observation but lacks 3D information.
RGB cameras, on the other hand, are generally more price-friendly and can capture high-resolution images rich in color and texture. However, they are fundamentally limited by their inability to capture 3D spatial information directly. This absence of information can pose challenges in determining the 3D pose of an object, limiting their applicability in most manipulation tasks.

\begin{figure}[t]\label{Teaser}
\begin{center}
    \includegraphics[width=\linewidth]{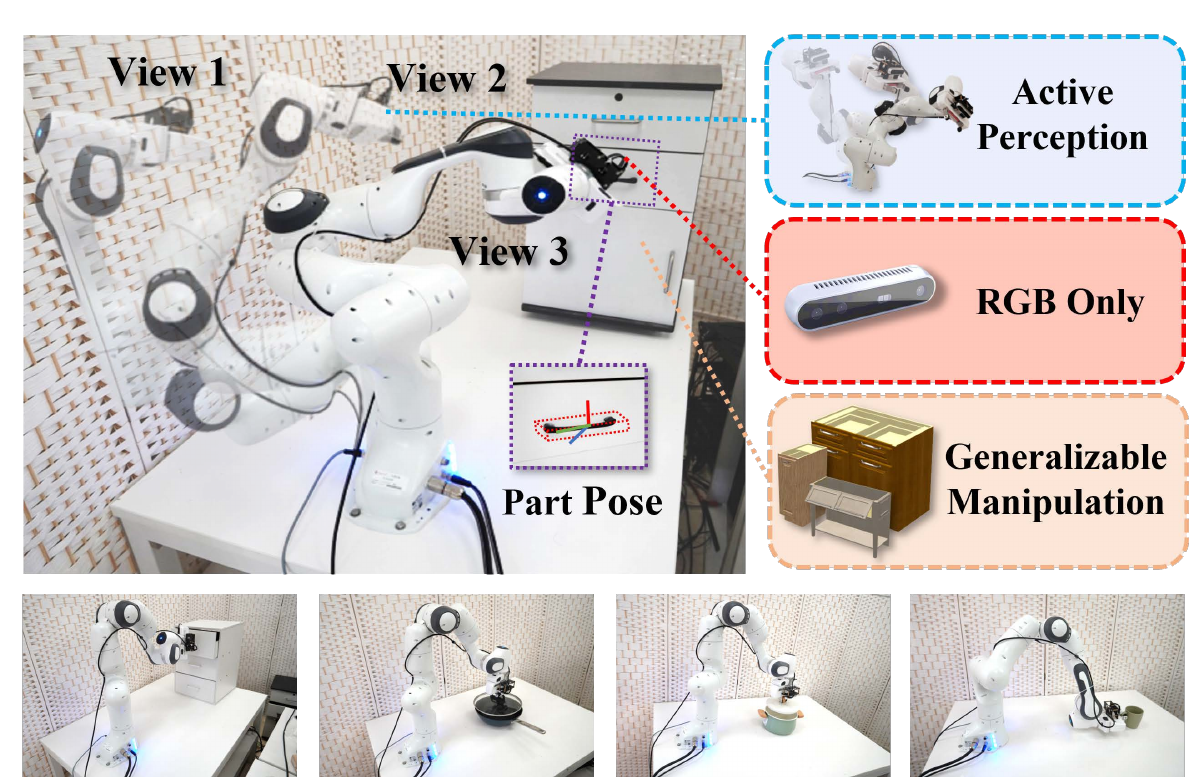}
\end{center}
\vspace{-0.3cm}
\caption{An eye-on-hand camera captures multiple RGB images to estimate the object pose in the manipulation process}
\vspace{-0.7cm}
\label{fig:teasor}
\end{figure}

To empower robots perceive the environment in a way rich in both high-resolution details and 3D information, and adjust adaptively according to the environment, we propose an image-only robotic manipulation policy that utilizes a single eye-on-hand camera to actively observe the environment to finish given manipulation tasks, as illustrated in Figure~\ref{Teaser}. Our approach decouples the manipulation process into three processes: 1) Global Scheduling, the first process proposes way points for the robot to explore the environment adaptively. Powered by reinforcement learning, this process enables the robot to adapt to different manipulation tasks (open door, pick mug, etc.) and gather information from different views, avoiding occlusions, making it possible for 3D representations in the next process. 
2) Active Perception, the second process takes as input RGB images from different viewpoints  that are captured while the gripper is approaching the object to be manipulated and learns to estimate the 6D pose of  either the entire object or a specific object part \emph{e.g.}, the pose of a mug on the table in the mug-picking task, or the pose of a door handle in the door-opening task. This process paves the road for the third process. 3) Manipulation, we use a control-based approach to manipulate the object given the pose estimation. A closed-loop impedance controller is used for higher reliability. The three processes are
% organized together through low-dimensional pose information of the target present in the environment, and
coordinated under the Global Scheduling process.

By decoupling the manipulation process into three different parts, our approach has several advantages over existing methods. First, it allows the robot to capture high-resolution visual information while also estimating the 6D pose of the object, which is crucial for accurate manipulation. Second, it enables the robot to adapt to different tasks and objects, enhancing its versatility and effectiveness.
% Finally, our approach is robust to changes in the environment and physical parameters, making it intrinsically well-suited for real-world robotic manipulation once trained in simulators.
Finally, our approach provides an option to balance accuracy and efficiency, solving the trade-off between exploration and exploitation.

% Additionally, we further proposed a dataset of real objects for real-world benchmarking, where the objects are accessible globally.

% Robotic manipulation requires accurate perception of the surrounding environment. Mainstream research usually uses RGB image or point clouds s input for environment perception. However, the two ways have their own drawbacks: RGB image lacks depth information which is critical for the robot to know the actual pose of an object while point clouds are always noisy and fails on specular and transparent objects which are very common in daily environment.

% In this work, we proposed a novel active sensing pipeline decoupling decision-making, visual perception and physical interaction process, achieving real-world applicability. 
% Based on the predicted part poses, we also proved that carefully designed rule-based closed loop force-feedback planning can outperform most learning-based control in category level manipulation both in simulators and in the real-world. Additionally, we further proposed a dataset of real objects making real-world benchmarking possible, objects in the dataset are accessible globally.

\section{Related Work}

\subsection{Vision-based Robotic Manipulation}

Human decision-making and locomotion heavily rely on visual perception. Similarly, visual perception plays a crucial role for robots to adapt to and interact with the real world. Recent years have witnessed significant advancements in vision-based robotic manipulation, where robots utilize visual information to perceive and understand their environment.

Various visual modalities have been explored for perceiving the environment in robotic manipulation. Some studies, such as Where2Act~\cite{mo2021where2act}, SAGCI-System~\cite{lv2022sagci}, RLAfford~\cite{geng2022end} and Flowbot3D~\cite{eisner2022flowbot3d}, have employed point clouds as observations, leveraging the 3D information they provide. On the other hand, approaches from Geng \textit{et al.} \cite{geng2023gapartnet}, Xu \textit{et al.} \cite{xu2022universal} and Wu \textit{et al.}~\cite{wu2020grasp} have utilized both RGB images and depth maps for tasks like articulated object manipulation and object grasping.
However, there has been limited exploration of using only RGB images as input, mainly due to the belief that depth information is essential for determining the actual 3D coordinates of pixels in an image, thus RGB-only input may result in spatial ambiguity. For instance, in the work Where2Act~\cite{mo2021where2act}, authors compared performance of policies with RGB image input and RGBD image input, and from the experimental result, we can see a large performance drop due to the removal of depth information. This highlights the trade-off between using RGB-only observations and the depth-rich RGBD input.
Nonetheless, the use of depth information poses challenges when dealing with specular and transparent textures, as they can interfere with the depth capturing process and result in noisy depth maps \cite{sajjan2020clear}. To address this issue, some works, such as \cite{ichnowski2021dex, dai2022graspnerf}, have employed Neural Radiance Field (NeRF) \cite{mildenhall2021nerf} to recover depth information from multi-view RGB images.

% Different visual modalities have been used in perceiving the environment. \cite{mo2021where2act,geng2022end,lv2022sagci,eisner2022flowbot3d} adopted point cloud as the observation, while \cite{xu2022universal,wu2020grasp} uses both RGB image and depth map for articulated object manipulation and object grasping. 
% However, few works have been using only RGB image as input because depth information is necessary for an algorithm to determine the actual 3D coordinate of any pixel in the image, thus lacking this modality results in spacial ambiguity. \cite{mo2021where2act} considered RGB-only policies, but comparing to the RGBD input policies in the original work, the performance dropped due to the loss of depth information. Once there is gain, there is also loss. \cite{sajjan2020clear} addressed some issue of using depth as input, that specular and transparent textures may interfere with current depth capturing process and result in a noisy depth map. \cite{ichnowski2021dex,dai2022graspnerf} tends to solve this problem by recovering depth information from multi-view RGB images via Neural Radiance Field (NeRF)\cite{mildenhall2021nerf}.

Despite the advancements in RGBD-based methods, our approach focuses on utilizing RGB images as the sole input modality for robotic manipulation.
% By avoiding the challenges associated with depth information, we circumvent the trade-off between RGB and depth observations.
Unlike NeRF-based approaches \cite{ichnowski2021dex, dai2022graspnerf}, our method does not rely on depth recovery. Instead, we directly estimate the 6D poses of objects using a kinematics-guided multi-view pose estimator. This unique approach allows us to circumvent the trade-off between RGB and depth observations.
Through our RGB-only approach with pose estimation, we contribute a novel perspective to the field of vision-based robotic manipulation, highlighting the potential of leveraging only RGB images for perception and control.
% This research direction opens up new possibilities for real-world-oriented manipulation research, where RGB-only observations can provide valuable insights and practical solutions.

\begin{figure*}[t]

    \begin{center}
        \includegraphics[width=\linewidth]{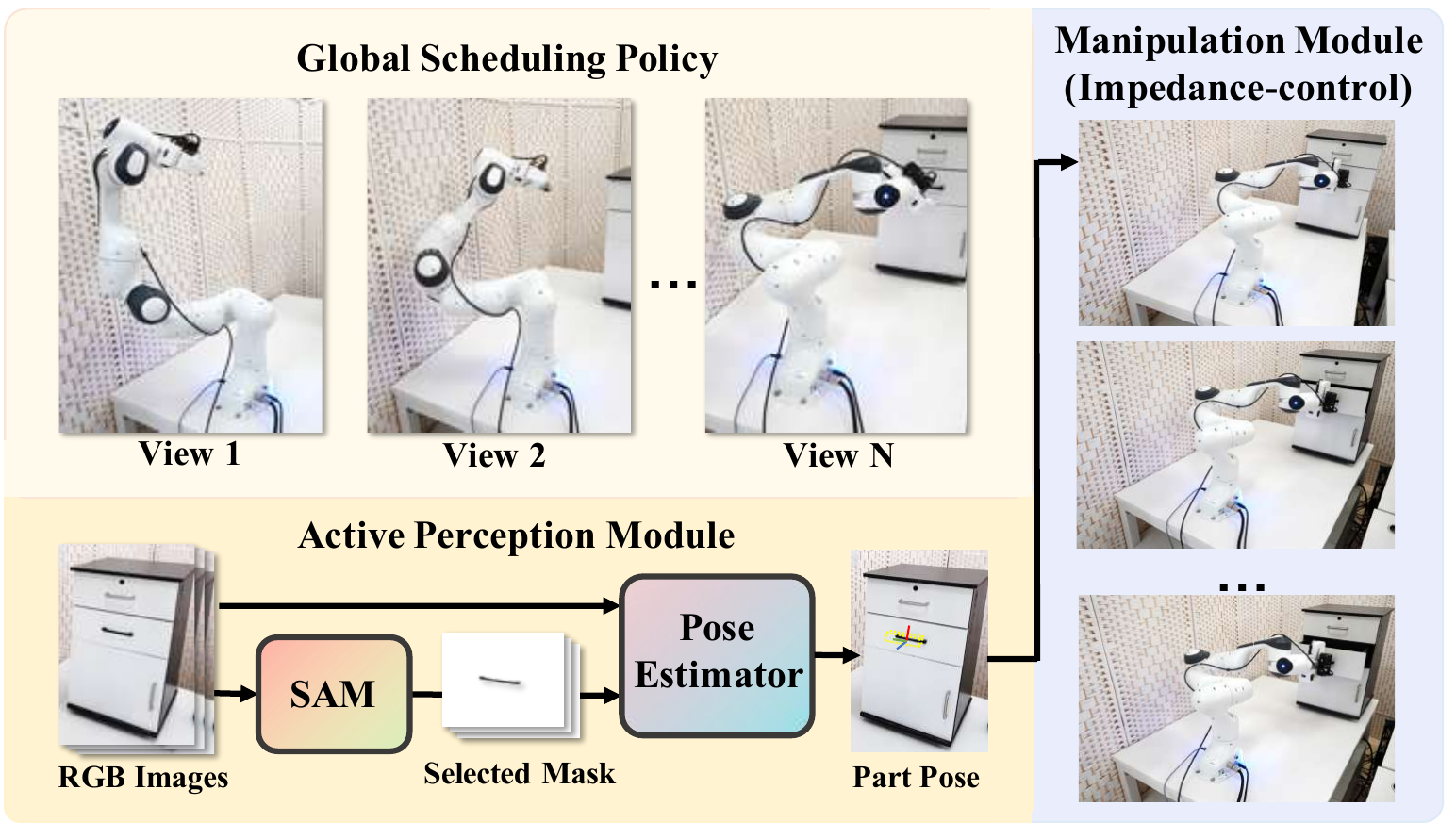}
    \end{center}
    \vspace{-0.3cm}
    \caption{In our pipeline, the Global Scheduling Policy serves as a high-level decision making policy to schedule Active Perception Module and Manipulation Module. The Active Perception Module learns to perceive the environment to predict pose information with the help of a pre-trained segmentation model (SAM~\cite{kirillov2023segment}). The Manipulation Module is used to complete the manipulation task through impedance control.
    }
    \vspace{-0.5cm}
    \label{fig:pipeline}
\end{figure*}
   
\subsection{Object Pose Estimation}
Object pose estimation provides position and orientation information for the target object, which are important for robotic manipulation.
In this work, we focus on category-level object pose estimation~\cite{sahin2018category}, which aims to predict the pose for unseen objects belonging to a specific object category.
Once the pose estimation model is trained, it can be directly applied to novel objects for robotic manipulation.
Most existing methods adopt a prior-based pose estimation paradigm~\cite{wang2019normalized,wang2021category}. 
Typically,
SPD~\cite{tian2020shape} optimizes a shape deformation field to reconstruct the 3D object model.
Then, it densely matches the reconstructed model and the observed instance point cloud for object pose estimation.
SGPA~\cite{chen2021sgpa} developes a prior adaptation module, which dynamically adjusts the prior feature to handle intra-class variation and achieves a higher category-level pose accuracy.
RBP-Pose~\cite{zhang2022rbp} further leverages a residual-vector-based representation to enhance the 3D spatial cues in the object point cloud for robust object pose estimation.
Recently, Liu \textit{et al.} proposed IST-Net~\cite{liu2023prior}, a prior-free framework.
It resorted to an implicit space transformation module, which associates the camera-space feature with the object-space feature in an implicit way without relying on any shape prior point cloud.
However, both prior-based and prior-free methods highly rely on object point clouds, which are not applicable when high-quality point cloud observation is not available.
To tackle this limitation, StereoPose~\cite{chen2023stereopose} proposed a pure image-based framework.
It leverages a parallax-aware module to fuse stereo image features and model intra-class object shape variation.
Stereo coordinate maps are further regressed from stereo images for accurate object pose estimation.
Inspired by StereoPose, in this work, we propose a novel multi-view image-based method for category-level object pose estimation.
Different from StereoPose, our method will recover object pose from images captured at multiple viewpoints along the robot trajectory.
To reduce the pose ambiguity of monocular images, 
we will leverage robot kinematics data to effectively fuse multi-view image features.
In addition, by actively adjusting the robot trajectory, we manage to utilize the most informative views to recover the object pose accurately.

% \subsection{Pipeline Structure}

\section{Method}

As shown in Fig.~\ref{fig:pipeline}, our method mainly consists of 3 modules, the Global Scheduling Policy $\mathrm S$, the Active Perception Module $\mathrm P$ and the Manipulation Module $\mathrm M$. The three modules are coordinated under the control of the Global Scheduling Policy $\mathrm S$.

Once the robot was deployed in the environment, it will explore the environment while completing the assigned manipulation task. In other word, the robot will actively perceive the surrounding environment through a camera mounted on its end-effector through the procedures described below.

\subsection{Exploration via Global Scheduling}
\label{GlobalScheduling}
% Global Scheduling Policy $\mathrm S$ is a high-level decision making policy powered by reinforcement learning algorithms. It will decide whether to explore the environment from a new view or to perform manipulation based on the information gathered and the results from the Active Perception Module. If it decides to explore from a new view, it also decides the 6D extrinsic of that view, and command the robotic arm to move to the position to take a picture. The picture will be used by Active Perception Module to provide pose estimations. If it decides to perform the manipulation task, then Manipulation Module will take over the control.
The Global Scheduling Policy, denoted as $\mathrm S$, serves as a high-level decision-making mechanism powered by reinforcement learning algorithms. Its primary role is to decide whether to further explore the environment from a novel perspective or to initiate manipulation, taking into account the accumulated information and the feedback from the Active Perception Module.
When opting for exploration, the policy specifies the 6D extrinsic parameters for the robotic arm to relocate to capture an image. This captured image is then utilized by the Active Perception Module to produce pose estimations.
Conversely, if the policy concludes that it's appropriate to stop the exploration, control will be transferred to the Manipulation Module.

% \subsection{Active Perception}

% Active Perception Module $\mathcal P$ generates pose information which the Global Scheduling Policy and Manipulation Module relies on. So this module greatly determines the overall performance of our pipeline.

% From a global view, The whole trajectory performed by the robot consists of several way-points $\mathrm p_1, \mathrm p_2,\cdots$ to explore the environment, and a sequence of motion to finish the manipulation. During the exploration, the robot's end-effector will follow the path determined by the way-points, and gather information from different views.

% save a view $\mathrm V_t$ from the RGB camera to a view buffer $\mathrm{VB}$ at the $t$-th way-point. Way-point $t$ is chosen by the Decision Making Module based on all views taken before $\mathrm V_t$.

More precisely, at time step $t$, the Global Scheduling Policy takes all previous views $\mathrm V_1,\cdots \mathrm V_{t-1}$ and the current prediction from Active Perception Module as input, and outputs two values: $\mathrm p_t$ and $f_t$ . The second output $f_t$ decides whether to terminate the view-point planning process and try to finish the manipulation task based on current information. If $f_t=0$, then the robot will continue exploration process and go to way-point $\mathrm p_t$ to obtain view $V_t$, otherwise, the Manipulation Module will take over the control of the robot. This modeling allows us to train the Global Scheduling Policy with Proximal Policy Optimization~\cite{schulman2017proximal}.

\subsection{Kinematics-Guided Multi-view Object Pose Estimation}
\label{Trick}
The core role of the Active Perception Module is to estimate the pose for the object of interest, given all information gathered during the exploration. 
At time step $t$, the camera mounted on the robot arm will capture an RGB image $I_t$ for the target object.
We exploited a segmentation model~\cite{kirillov2023segment} to crop the object region, as shown in Fig.~\ref{fig:pipeline}.
In order to handle the intra-class variation for category-level object pose estimation, similar to~\cite{chen2021sgpa,zhang2022rbp,liu2023prior}, we first resorted to a canonical object representation~\cite{wang2019normalized} and estimated a normalized coordinate map based on $F_t$, which is the deep feature of $I_t$.
The predicted coordinate map $M_t$ encodes dense 2D-3D correspondences between the camera and object frame, which are essential for object pose estimation.
However, the category-level pose cannot be fully recovered with monocular RGB image, due to the depth ambiguity.
In this regard, we further proposed a kinematics-guided depth-aware module to fuse multi-view image features.
It aims to leverage the robot kinematics data to reduce the pose estimation ambiguity.
For adjacent two RGB images $I_t$ and $I_{t+1}$, their relative extrinsic $(\mathbf{R}_{t}^{t+1}, \mathbf{t}_{t}^{t+1})$ can be derived from the kinematics data between $t$ and $t+1$.
Then, we fused adjacent image features by warping $F_t$ to $F_{t+1}$ with a multi-homography mapping.
Specifically, we uniformly sample a set of hypothetical depth planes $\{d_i\}_{i=1}^N$ between $d_{min}$ and $d_{max}$.
At each hypothetical depth plane, we warp $F_t$ to $F_{t+1}$ based on the corresponding homography, which is computed as:
\begin{equation}
    H(d_i)=\mathbf{K}\cdot\mathbf{R}_{t}^{t+1}(\mathbf{I}+\frac{\mathbf{t}_{t}^{t+1}\cdot\mathbf{n}^\top}{d_i})\cdot\mathbf{K}^{-1},
\end{equation}
where $\mathbf{K}$ denotes the camera intrinsics and $\mathbf{n}$ denotes the principle axis of the camera at time step $t+1$.
The warped features would exhibit different similarities on different depth planes.
Therefore, by concatenating features at different depth, we can construct a 4D depth-aware feature volume. This volume is then regularized with a volume regularization layer similar to~\cite{yao2018mvsnet,chen2021mvsnerf} to derive the fused image feature $\hat{F}_{t+1}$. $\hat{F}_{t+1}$ is further concatenated with the features extracted from $M_{t+1}$ and passed through MLP-based networks to predict object size, rotation and translation, respectively.
% We proposed an active multi-view pose estimator to achieve accurate prediction of 6-D pose of critical parts of the objects in the environment. The estimator is task-specific, which each task uses a separately trained model.
% \info{CK}

\begin{figure}[h]
    \centering
    \includegraphics[width=1.0\linewidth]{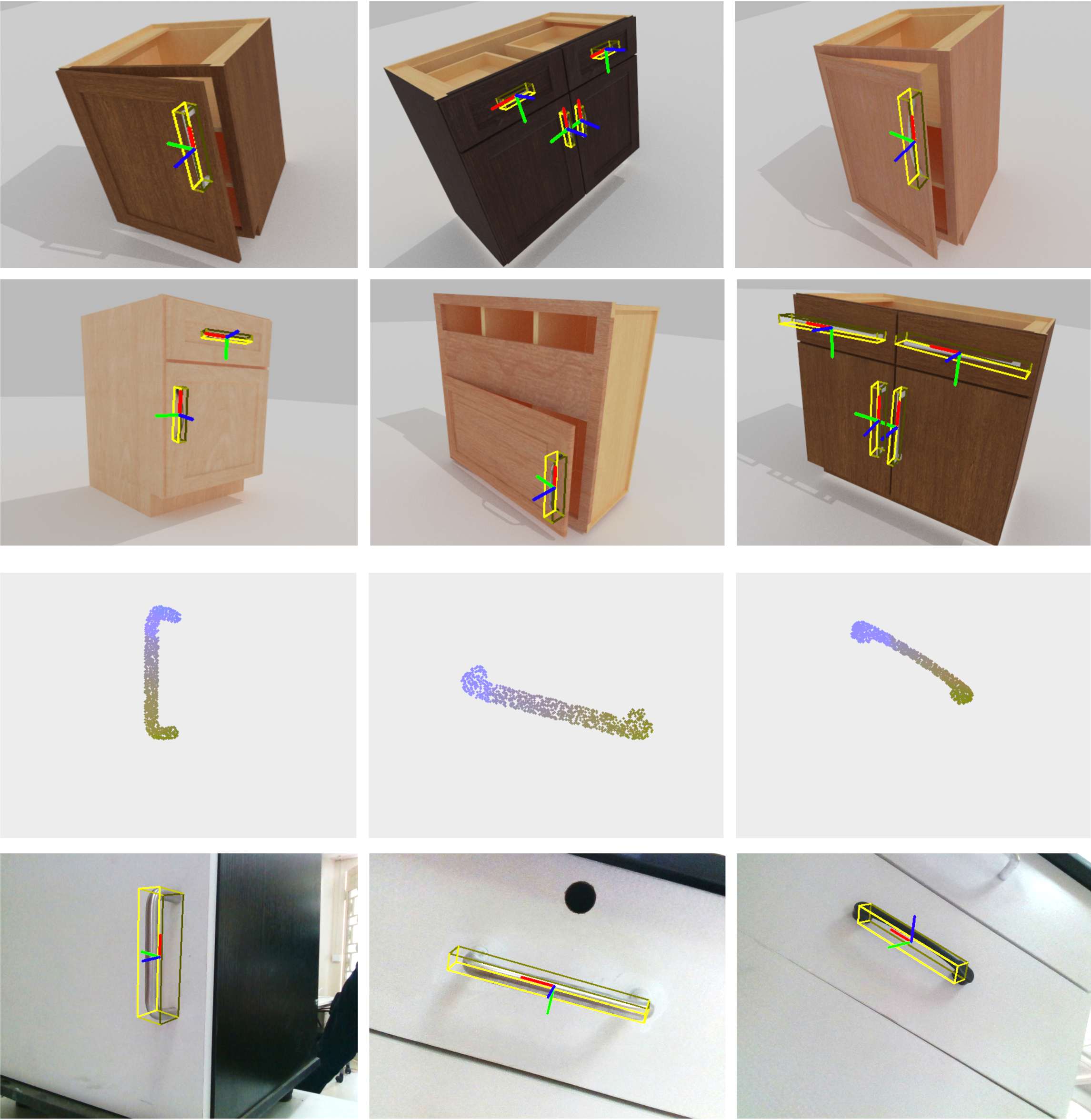}\vspace{-5mm}
    \caption{Category-level object pose estimation results for handles of different cabinets in the simulator~(the first two rows) and the real world~(the bottom two rows).}
    \vspace{-0.5cm}
    \label{fig:pose-result}
\end{figure}

\subsection{Domain Randomization}
\label{DR}
To better adapt to real-world scenarios, we added domain randomization to the training environment. While training the active perception module, the texture of objects (including transparent, specular and diffuse material, which presents challenges for point-cloud cameras but not for RGB cameras), the position and intensity of light source and the initial pose of the object will change.

\subsection{Balancing Accuracy and Efficiency (A \& E)}
\label{Balancing}
% In training of Decision Making Module, we incorporated both the success rate of the whole task and the time it takes to finish the task, forcing it to balance between accuracy and efficiency. More accurate pose estimation will increase the success rate of the whole task, but needs more views and the distance between adjacent way-points will be larger. Similarly, increasing the efficiency requires fewer way-points, affecting the accuracy and success rate.
%% refine by ChatGPT:
While the balance of exploration and exploitation is a fundamental problem in policy learning~\cite{ladosz2022exploration,amin2021survey}, there is a similar problem which we name it the balance of accuracy and efficiency.
%During the training of the Active Perception Module, we incorporated two performance metrics: the precision of pose estimation and the time it takes to complete the task.  This requires a balance between accuracy and efficiency to be made.
Higher accuracy of pose estimation may enhance the success rate of the entire task, but usually requires more views and increases the total distance between way-points, harming the efficiency of the method. On the other hand, increasing the efficiency necessitates fewer way-points and less variety of view points, negatively impacts the estimation accuracy and success rate.
%Therefore, by considering both the precision of pose estimation and the time required to complete the task, we can effectively illustrate the trade-off between accuracy and efficiency.

To consider both the precision and efficiency in this balance, we introduced a parameter $\alpha$ in the reward computation of the Active Perception Module. This parameter is defined as $\alpha=\frac{r_{pen}}{r_{prec}}$ , where $r_{prec}$ is the reward for the precision of pose-estimation and $r_{pen}$ is the penalty for moving distance. A smaller $\alpha$ biases the system towards better precision, while larger $\alpha$ tends to minimize the time required.

\subsection{Impedance-control Manipulator}
\label{Impedance}
Using visual-aware closed-loop control in the context of active perception is less favored because it often results in inadequate visual information to estimate crucial states. For instance, when our monocular robot opens a door using the door handle, the camera is positioned too closely, making it challenging to observe the door rotation.

To overcome this limitation, we turned our attention to harnessing more information from robotic kinematics. The force exerted on the robotic arm can be used as a signal to adjust the manipulation. Our manipulator employs an impedance controller. Here, the end-effector has the freedom of movement but tends to return to its target pose, ensures its tolerance to minor errors. Given $\mathbf X$ and $\mathbf R$ as the translational and rotational error of the end-effector from its target pose, the torque $\mathbf \tau$ for each robot joint is computed as:
\begin{equation}
    \mathbf\tau = \mathbf J^T\left(
    -k
    \begin{pmatrix}
    \mathbf X\\ \mathbf R
    \end{pmatrix}
    -b (\mathbf J\mathbf{\dot q})
    \right) + \mathbf N,
\end{equation}
Where $\mathbf J$ denotes the Jacobian matrix of the robot, $k,b$ represent the stiffness and damping terms, respetively. The variable $\mathbf q$ is the current robot joint state, and $\mathbf N$ is the additional term account for handle Jacobian nullspace and Coriolis force.
Viewing the manipulation trajectory as a time-dependent function, we can dynamically predict the subsequent point on the trajectory, leading to a reliable manipulation policy that remains resilient to disturbances and can effectively manage both revolute and prismatic articulated objects. Let $\mathbf{p}$ denote the pose of the end-effector over time. Then we determine the current target pose as:
\begin{equation}
    \mathbf{p}^*=\mathbf{p}+k_1 \mathbf{\dot p}+k_2 \mathbf{\ddot p},
\end{equation}
Here $k_1$ and $k_2$ are coefficients for correcting direction and curvature of the trajectory.

\section{Experiment}

\subsection{Task Settings}
We designed six challenging tasks to evaluate our method. In all tasks, a robotic arm is required to accomplish a specific manipulation goal with different objects.

\textbf{Open Door: } A door is initially closed, the agent needs to open the door larger than 0.15 rad (8.6 degrees). The position and rotation of the door is randomized within a range to make the task more challenging.

\textbf{Open Door 45\si{\degree}: } The harder version of Open Door. The agent needs to open the door to more than 45 degrees.

\textbf{Open Drawer: } A drawer is initially closed, the agent needs to open the drawer larger than a specific distance (15 centimeters). The position and rotation of the drawer is also randomized.

\textbf{Open Drawer 30cm: } The harder version of Open Drawer. The agent needs to open the drawer to more than 30 centimeters, which is fully open.

\textbf{Open Pot: } A kitchen-pot is initially on the floor with its lids on, the agent needs to lift the lid to a specific height. The position and rotation of the pot is also randomized.

\textbf{Pick Mug: } A mug is initially on the floor, the agent needs to pick up it to a specific height. The position and rotation of the mug is also randomized.

% \subsection{Dataset}
To evaluate our method, we trained our models using two datasets: PartNetMobility, a 3D articulated object dataset~\cite{Mo_2019_CVPR}, and ShapeNet, a comprehensive rigid 3D shape dataset~\cite{chang2015shapenet}. All training was conducted within the SAPIEN simulator~\cite{Xiang_2020_SAPIEN, gu2023maniskill2}. Within the simulator, our experiments spanned 184 shapes from 4 distinct object categories. Additionally, we selected real-world objects for testing.
% that consists of daily objects from international brands including MUJI\footnote{https://www.muji.us/} and IKEA\footnote{https://www.ikea.com/us/en/}, making it possible for real-world benchmarking. The item list of both the simulated and real-world dataset can be found in the appendix.
%, and one real-world 3D cabinet scan we capture using a ZED MINI RGB-D camera.

% \subsection{Quantitaive Evaluation}

\begin{table*}[h]
\setlength\tabcolsep{5.4pt}
% \caption{\textbf{Quantitative results with modalities.}\\ Success Rate (\%)}
\caption{Quantitative comparison between our method and baselines} \vspace{-0.4cm}   
\label{table1}
\center
\begin{tabular}{c|c|cc|cc|cc|cc|cc|cc}
\hline
\multirow{2}{*}{\textbf{Methods}}
    & \multirow{2}{*}{\textbf{Modality}} & \multicolumn{2}{c|}{\ \ \textbf{Open Door}\ \ }  & \multicolumn{2}{c|}{\ \ \textbf{Open Door 45\si{\degree}}\ \ \ }                        & \multicolumn{2}{c|}{\ \ \textbf{Open Drawer}\ \ } & \multicolumn{2}{c|}{\ \ \textbf{Open Drawer 30cm}\ \ }                      & \multicolumn{2}{c|}{\ \ \textbf{Open Pot}\ \ }                         & \multicolumn{2}{c}{\ \ \textbf{Lift Mug}\ \ }                         \\
    &                                      & \multicolumn{1}{c}{Train} & \multicolumn{1}{c|}{Test} & \multicolumn{1}{c}{Train} & \multicolumn{1}{c|}{Test} & \multicolumn{1}{c}{Train} & \multicolumn{1}{c|}{Test} & \multicolumn{1}{c}{Train} & \multicolumn{1}{c|}{Test} & \multicolumn{1}{c}{Train} & \multicolumn{1}{c|}{Test} & \multicolumn{1}{c}{Train} & \multicolumn{1}{c}{Test} \\
    \hline
Where2Act & Point-clouds & 8.0 & 7.0 & 1.8 & 2.0 & 5.9 & 7.5 & 1.1 & 0.6 & \textbf{30.0} & 55.3 & 20.9 & 19.6 \\

Flowbot3D & Point-clouds & 19.5 & 20.4 & 6.8 & 6.4 & 27.3 & 25.8 & 16.9 & 11.3 & 2.5 & 7.4 & 4.9 & 4.3\\
                                    
UMPNet & Point-clouds & 27.1 & 28.1 & 11.0 & 10.9& 16.6 & 18.8 & 4.4 & 5.6 & 19.1 & 36.9 & 26.6 & 22.9                       \\

GAPartNet & Point-clouds & 69.5 & 74.5 & 39.4 & 43.6 & 50.6 & 59.3 & 44.6 & 48.6 & 5.3 & 10.8 & 0.0 & 0.0 \\

\hdashline[2pt/2pt]

DrQ-v2 & RGB & 1.8 & 2.5 & 0.8 & 0.8 & 1.9 & 1.0 & 1.4 & 0.5 & 0.1 & 0.0 & 0.0 & 0.0 \\
LookCloser & RGB & 1.5 & 1.25 & 0.8 & 0.8 & 0.8 & 0.0 & 0.0 & 0.0 & 0.3 & 0.0 & 4.8 & 6.5 \\
                          
                                     % \hdashline[1pt/1pt]
\textbf{Ours} & RGB & \textbf{89.3} & \textbf{88.9} & \textbf{47.8} & \textbf{52.9} & \textbf{83.0} & \textbf{87.0} & \textbf{63.5} & \textbf{61.9} & 22.8 & \textbf{55.6} & \textbf{48.4} & \textbf{41.9} \\
                                   \hline

\end{tabular}
% \vspace{-0.2cm}
\end{table*}

\begin{table*}[h]
\setlength\tabcolsep{5.4pt}
\caption{%Average success rate and master percentage (ASR/MP) 
Ablation Study of Our Method}%\\ Success Rate (\%)}
\vspace{-0.4cm}   
\label{table2}
\center
\begin{tabular}{c|cc|cc|cc|cc}
\hline

\multirow{2}{*}{\textbf{Methods}} & \multicolumn{2}{c|}{\ \ \textbf{Open Door}\ \ } & \multicolumn{2}{c|}{\ \ \textbf{Open Door 45\si{\degree}}\ \ } & \multicolumn{2}{c|}{\ \ \textbf{Open Drawer}\ \ } & \multicolumn{2}{c}{\ \ \textbf{Open Drawer 30cm}\ \ } \\
% \cmidrule{r}{6-8}
& \multicolumn{1}{c}{Train} & \multicolumn{1}{c|}{Test} & \multicolumn{1}{c}{Train} & \multicolumn{1}{c|}{Test} & \multicolumn{1}{c}{Train} & \multicolumn{1}{c|}{Test} & \multicolumn{1}{c}{Train} & \multicolumn{1}{c}{Test} 
\\
\hline
% Ours w/o Global Scheduling                          &  &  &  &  &  & &  &  \\               
Ours w/o Global Scheduling & 78.6 & 78.4 & 49.8 & 39.1 & 82.4 & 83.3 & \textbf{64.8} & 56.6 \\
Ours w/o Impedance Control & 74.5 & 74.0 & 24.6 & 29.3 & 68.9 & 68.4 & 24.6 & 29.3\\
Ours w/o Domain Randomization                       & 66.6 & 73.0 & 32.5 & 36.4 & 77.9 & 77.6 & 40.6 & 30.1 \\
Ours w/o Pose Estimation Tricks                     & 65.5 & 36.0 & 48.3 & 22.8 & 35.9 & 42.0 & 43.0 & 20.3 \\
% Ours w/o Balancing A\&E                           &                           &                            \\
\textbf{Ours} & \textbf{89.3} & \textbf{88.9} & \textbf{51.1} & \textbf{52.9} & \textbf{83.0} & \textbf{87.0} & 63.5 & \textbf{61.9} \\
\hline
\end{tabular}
\vspace{-0.5cm}
\end{table*}

For each task, we divided the objects into a training set and a testing set, trained our method, baselines and ablations fully on the training set and saved checkpoints every 25 time-steps within 2000 total time-steps. Then, we selected the checkpoint with the highest reward for comparison. We use average success rate to evaluate our method.

% The criteria for success is the displacement for the articulated joint greater than a specific limit. For example, for the Open Door task, we require an opening of 54 degrees.

% \textbf{Operation Time}: We measure the average time consumption of a policy finishing the manipulation task over the whole dataset. This number indicates the efficiency of the policy.

\begin{figure}[t]
    % #of views v.s. success rate
    \begin{center}
        \includegraphics[width=\linewidth]{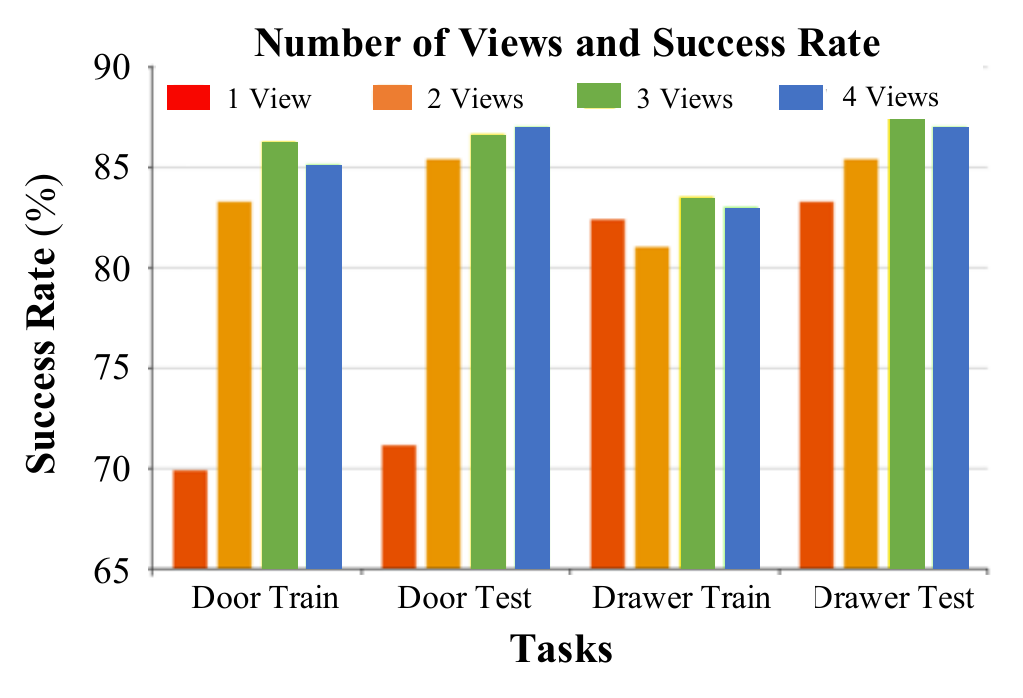}
    \end{center}
    \vspace{-0.5cm}
    \caption{Performance of our method under different number of views.}
    \vspace{-0.5cm}
    \label{fig:balancing_viewpoint}
\end{figure}

\begin{figure}[t]
    % distance v.s. success rate
    \begin{center}
        \includegraphics[width=\linewidth]{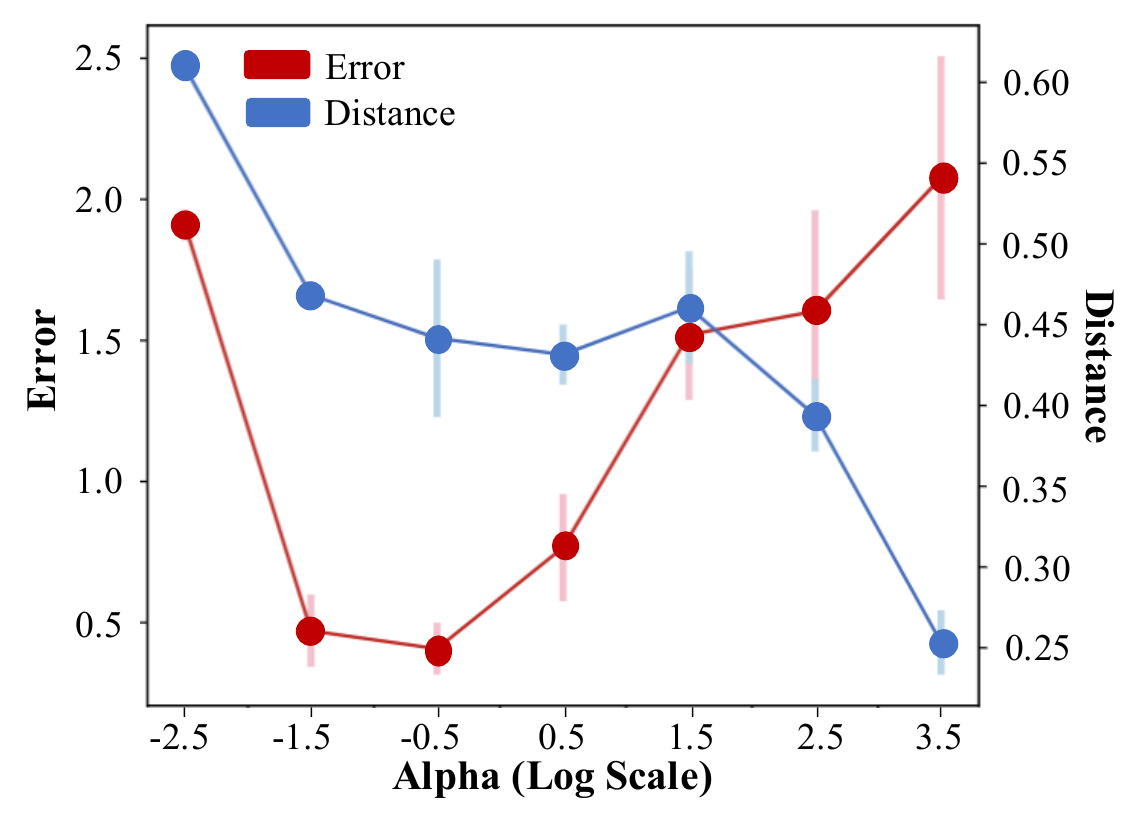}
    \end{center}
    \vspace{-0.5cm}
    \caption{Our method under different $\alpha$ for balancing A\&E. The x-axis indicates the value of $\alpha$, the red curve corresponds the error of pose estimation and the blue curve is the average moving distance during manipulation. The dot and bar are the mean and standard deviation (respectively) over 5 differently evaluated policies.}
    \vspace{-0.5cm}
    \label{fig:balancing_alpha}
\end{figure}

\subsection{Baselines and Ablation}
We benchmarked our method against six other algorithms, categorizing them into two groups based on their input type: four point-cloud-based and two that exclusively use RGB. The results are summarized in Table.~\ref{table1}. The following is a brief description of each:

\textbf{Where2Act~\cite{mo2021where2act}}: Operates on point-cloud inputs, estimating per-point action scores. To execute the task, we integrated it with our manipulation policy, selecting the point with the highest score for interaction.

\textbf{Flowbot3D~\cite{eisner2022flowbot3d}}: Predicts the point-wise motion direction on the point cloud, denoting it as 'flow'. The point with the largest flow magnitude serves as our interaction point. Subsequent manipulations utilize our policy. Notably, we replaced the original suction gripper with a parallel one to ensure a fair comparison.

\textbf{UMPNet~\cite{xu2022universal}}: Accepts RGBD images, predicting an action point on the image which is then projected into 3D space based on the depth data of the predicted pixel. Similar to the above methods, we paired it with our manipulation policy. The original suction gripper in this method was also replaced with a parallel gripper for comparison.

\textbf{GAPartNet~\cite{geng2023gapartnet}}: A pose-centric approach that predicts the pose of an object's part from point-cloud inputs. Manipulations are executed based on the predicted pose of the relevant task's component using our policy.

\textbf{DrQ-v2~\cite{yarats2021mastering}}: Represents the cutting-edge in pure RL methodologies. Here, reinforcement learning directly trains the manipulation policy. Inputs for this policy encompass both the robot's state and an RGB image, culminating in an output specifying the desired 6D pose of the robot's end-effector.

\textbf{LookCloser~\cite{jangir2022look}}: A multi-view RL model combining third-person and egocentric viewpoints. While DrQ-v2 is confined to a single eye-on-hand camera's image input, LookCloser's used of multi-view input and visual transformers~\cite{dosovitskiy2020transformers} enables the fusion of data from varied angles.

To elucidate the contribution and effectiveness of individual modules within our approach, we conducted an extensive ablation study. Six experiments were carried out, each omitting or adjusting specific components:

\textbf{Ours w/o Global Scheduling}: Rather than leveraging the observation perspective determined by the Global Scheduling Policy (Sec~\ref{GlobalScheduling}), this experiment uses two manually set fixed perspectives for perception.

\textbf{Ours w/o Impedance Control}: This variant employs an open-loop manipulation policy. In the absence of the impedance control manipulator (Sec~\ref{Impedance}), the policy operates by moving directly to the desired position.

\textbf{Ours w/o Domain Randomization}: We trained our method without employing the domain randomization process outlined in Sec~\ref{DR}.

\textbf{Ours w/o Pose Estimation Tricks}: This experiment omits the kinematics-guided depth-aware fusion module from the object pose estimator, as detailed in Sec~\ref{Trick}.

\textbf{Ours w/ Different Number of Views}: This set of tests alters the approach by varying the number of views. The results underscore the diminishing returns of adding extra viewpoints, emphasizing the importance of striking a balance between accuracy and time-efficiency.

\textbf{Ours w/ Balancing A\&E}: For this group, the number of views is held constant at four. By adjusting the parameter $\alpha$ in the reward computation of the Global Scheduling Policy, this experiment showcases the interplay between accuracy and efficiency.

\subsection{Quantitative Results in Simulator} % 考虑下，要不要加个讨论，讲讲为什么相互帮助，为什么briding perceptron和manipulation了。放到 method作为 subsection也行。
% need modify, this section is a copy of rlafford
Table.~\ref{table1} shows our large-scale evaluation in simulator over different tasks.
%the results of point-cloud based baselines are significantly higher than RGB-only baselines, indicating the difficulty of obtaining 3D information from RGB images.
The results indicate that our method consistently surpasses all baselines across nearly all tasks. Notably, in the Open Pot task, the majority of methods exhibit improved performance on the test set. This observation can be attributed to the train-test split used by all methods, with the test set encompassing a greater number of simpler instances. It's also worth highlighting the significant performance decline of GAPartNet in the Pot and Mug tasks. This is likely a consequence of substantial pose estimation errors, especially considering the heightened precision required to pinpoint critical parts of the objects in these tasks.

As depicted in Table~\ref{table2}, each module within our proposed methodology plays a pivotal role. The effects of omitting the impedance control become increasingly pronounced in more intricate tasks (Open Door 45\si{\degree} and Open Drawer 30cm). This underscores the indispensability of the closed-loop impedance controller, particularly in long-horizon manipulative tasks.

Fig.~\ref{fig:balancing_viewpoint} reveals an intriguing trend: the augmentation of way-points directly correlates with an elevation in the average success rate of manipulations. However, the addition of more way-points inevitably leads to diminishing returns. With up to 4 way-points, no significant improvement is observable. These results reinforce the importance of finding a harmony between accuracy and efficiency.

Lastly, Fig.~\ref{fig:balancing_alpha} illustrates that the pose estimation accuracy diminishes when $\alpha$ is excessively large. Concurrently, the average moving distance experiences a decline as $\alpha$ increases. The unintuitive U-shaped error curve is possibly due to imperfect reward design, which the terms other than error and distance penalty dominates the overall reward when $\alpha$ is too small, leading to a sub-optimal policy.
% This pattern signifies that the equilibrium between accuracy and efficiency can be modulated by adjusting $\alpha$.

\subsection{Real-world Experiment}

% \begin{figure}[h]
%     \begin{center}
%         % \includegraphics[scale=0.5, 
%         \includegraphics[width=\linewidth]{fig/real.png}
%     \end{center}
%     \caption{Ten test objects of our real-world experiment.}
%     \label{fig:real-world-exp}
% \end{figure}

% In our real-world experiments, the agent independently explores the environment and successfully completes the assigned tasks. We tested the policies trained in the simulator using real objects from our real-world dataset.
We employed a Franka Panda robot arm as our manipulator and fixed a Realsense camera with RGB-only output onto the robot's end-effector. The observations from the camera are directly used by the agent without refinement. %Due to safety concerns, a human inspector will be present to oversee the execution of the robot, who will decide whether an action should or should not be interrupted, but will not change the observation of the action. 
Videos can be found on \href{https://rgbmanip.github.io/}{https://rgbmanip.github.io/}.

\section{Conclusion}

%We introduce the first work to explore active pose estimation for monocular robotics manipulation. Our method enables a robot to manipulate material-agnostic articulated objects, by first exploring the environment actively, then generate pose information of task-relevant objects, finally complete the manipulation task according to the pose information with a force-aware impedance control policy.
In this study, we presented a pioneering approach to active pose estimation for monocular robotic manipulation. Our method uniquely equips robots with the ability to handle different tasks with monocular RGB inputs. This is achieved through a three-pronged process. 1) The robot explores the environment actively. 2) Pose information of interested objects is derived from the exploration. 3) Manipulation is achieved with a closed-loop impedance control policy.

% The significance is that we achieve robust manipulation control, without using point-cloud sensors. Experimental results show that our method outperforms all baselines. %without significant performance drop on transparent and specular objects.

A notable implication of our work is the attainment of robust manipulation control without the necessity for point-cloud sensors. Furthermore, experimental evaluations solidify the superiority of our method, as it consistently surpassed all baseline approaches.

\section*{ACKNOWLEDGEMENT} This project was supported by The National Youth Talent Support Program (8200800081) and National Natural Science Foundation of China (No. 62136001).
This work was partially supported by a grant from the Research Grants Council of the Hong Kong Special Administrative Region, China (Project Number:  24209223).

\bibliographystyle{plain}
\bibliography{reference}
%%%%%%%%%%%%%%%%%%%%%%%%%%%%%%%%%%%%%%%%%%%%%%%%%%%%%%%%%%%%

\section{Appendix}

\subsection{Dataset}

We used 3D models from PartNetMobility~\cite{Mo_2019_CVPR} and ShapeNet~\cite{chang2015shapenet}. Details can be found on \url{https://rgbmanip.github.io}.

\subsection{Training Details}

\subsubsection{Global Scheduling Policy}

The Global Scheduling Policy solves the scheduling problem as a Markov Decision Process. We used Proximal Policy Optimization (PPO) to train our Global Scheduling Policy. The reward function is the weighted sum of the following terms:

\begin{itemize}
    \item Move-target difference reward: $\left\|p_{cam}-p_{tar}\right\|_2$, where $p_{cam}$ is the current position of the camera. $p_{tar}$ is current target position of the camera.
    \item Move success reward: $\mathbb{I}(\text{Move success})$
    \item Move period penalty: the number of steps used to move from previous position to the new one.
    \item Distance penalty: $\left\|p_{cam}-p_{prop}\right\|$, where $p_{prop}$ is a point $0.9m$ above the base of robot.
    \item Orientation reward: $q_{cam}\cdot q_{prop}$, $q_{cam}$ is the quaternion which the camera can face directly to the object.
    \item Look-at regularization penalty: $\left(\left\|p_{look\ at}-p_{tar}\right\|_2-1\right)^2$, where $p_{look\ at}$ is the target position where the camera should be facing at.
    \item Mask bounding-box penalty: $\left\|mid-[0.5\ 0.5]^T\right\|_2$, $mid$ is the coordinate of the central pixel of the bounding box of the object in the current view.
    \item Mask bounding-box boundary penalty: $\mathbb{I}(l\leq 0.1) + \mathbb{I}(r\geq 0.9) + \mathbb{I}(d\leq 0.1) + \mathbb{I}(u\geq 0.9)$, where $l,r,d,u$ are the boundaries of the bounding box.
    \item Object-in-view reward: $\mathbb{I}(\text{object in current view})$
    \item Center reward: $\frac{1}{1+\left\|p_{pred}-p_{gt}\right\|_2^2}$, here $p_{pred}$ and $p_{gt}$ are the predicted and ground-truth center of the object.
    \item Orientation reward: $\frac{1}{1+\left\|o_{pred}-o_{gt}\right\|_2^2}$, here $o_{pred}$ and $o_{gt}$ are the predicted and ground-truth orientation of the object.
    \item View diversity reward: $\mathbb{I}(\langle p_{cam}-p_{obj}, p_{cam}'-p_{obj}\rangle>0.3)$.
\end{itemize}

The model itself consists of a policy network and a value network. Each network uses a separate MLP with hidden layer size [96, 96, 32]. We used adaptive learning rate varying from $2\times 10^{-4}$ to $5\times 10^{-3}$ to train this model.

\subsubsection{Multi-view Object Pose Estimator}

The object pose estimation model was trained ahead of the Global Scheduling Policy based on pre-collected synthetic data.
We captured multi-view images in the simulator at viewpoints that were uniformly sampled from the hemisphere around the target object.
Then, random image pairs were fed into the pose estimator for model training.
For the homography-based feature fusion, we sampled the hypothetical depth plane between 0.1 and 2.4 with an interval of 0.1.
For each object part, the model predicted its normalized coordinate map, depth map, and object pose and size parameters. 
The object rotation was parameterized with a continuous 6D representation~\cite{zhou2019continuity}.
The training loss is the weighted sum of the following terms:
\begin{itemize}
    \item Pose loss: $\|\mathbf{R}_{pred}-\mathbf{R}_{gt}\|_2$ + $\|\mathbf{t}_{pred}-\mathbf{t}_{gt}\|_2$ + $\|\mathbf{s}_{pred}-\mathbf{s}_{gt}\|_2$, where $(\mathbf{R,t,s})_{pred}$ and $(\mathbf{R,t,s})_{gt}$ are the predicted and ground-truth pose parameters.
    \item Coordinate map loss: $\|C_{pred}-C_{gt}\|_1$, where $C_{pred}$ and $C_{gt}$ are predicted and ground-truth coordinate maps respectively.
    \item Depth loss: $\|D_{pred}-D_{gt}\|_1$, where $C_{pred}$ and $C_{gt}$ are predicted and ground-truth depth maps respectively.
\end{itemize}
For different object categories, the object pose estimation model was trained separately.

\subsubsection{Impedance-control Manipulator}

This model does not require training.

\subsection{Environment Settings}

In preparation for the setup of our real-world experimental environment, we assembled a collection of fifteen test objects, distributed across three categories: mugs, cabinets, and pots, with each category comprising five distinct varieties.

\subsubsection{Simulation}
The Franka robotic arm was designated as our agent, onto which an RGB camera was mounted at the gripper's location. An object is located in front of the robotic arm as the manipulation target. For the observation of our agent, a mask for the object is captured along with the RGB image from the mounted camera.

\subsubsection{Real-world}
Our operational pipeline in the real-world setting closely paralleled that of the simulator, with a singular distinction being our approach to mask selection. This process was made more flexible and observable by allowing mask selection to be guided either by prompts from the SAM model or through manual annotation.

\subsubsection{Domain-randomization for pose estimation} We employed the Sapien~\cite{Xiang_2020_SAPIEN} rendering engine to advance texture and lighting randomization, aiming to improve synthetic dataset realism for our tasks. The randomization includes different materials (transparent, specular and diffuse) of the same object, the intensity (strong or weak) and location (sampled on vertices and edge-centers on a surrounding cube) of light source will also change. This randomization process enriches our dataset for the pose estimator with varied appearances and lighting.

The initial pose of the target object is selected randomly. More precisely, the pose can be designated as a tuple of four values $(\alpha, \beta, d, h)$. $\alpha$ is the rotation along the z-axis of the object, $\beta$ is the azimuth relative to the robotic arm, $d$ is the distance from the robotic arm and $h$ is the height of the object. For different tasks, those values has different distributions. The unit of angles are radians in Table.\ref{table3}.

\begin{table}[h]
\setlength\tabcolsep{5.4pt}
\caption{Distribution of Parameters}
\vspace{-0.4cm}   
\label{table3}
\center
\resizebox{\linewidth}{!}{
\begin{tabular}{c|cc|cc|cc|cc}
\hline

\multirow{2}{*}{\textbf{Tasks}} & \multicolumn{2}{c|}{\textbf{$\alpha$}} & \multicolumn{2}{c|}{\textbf{$\beta$}} & \multicolumn{2}{c|}{\textbf{$d$}} & \multicolumn{2}{c}{\textbf{$h$}} \\
% \cmidrule{r}{6-8}
& \multicolumn{1}{c}{Low} & \multicolumn{1}{c|}{High} & \multicolumn{1}{c}{Low} & \multicolumn{1}{c|}{High} & \multicolumn{1}{c}{Low} & \multicolumn{1}{c|}{High} & \multicolumn{1}{c}{Low} & \multicolumn{1}{c}{High} 
\\
\hline
% Ours w/o Global Scheduling                          &  &  &  &  &  & &  &  \\               
Open Door & -0.20 & 0.20 & -0.40 & 0.40 & 0.50 & 0.85 & 0.01 & 0.05 \\
Open Drawer & -0.20 & 0.20 & -0.40 & 0.40 & 0.50 & 0.80 & 0.01 & 0.05 \\
Open Pot & -0.20 & 0.20 & -0.40 & 0.40 & 0.20 & 0.38 & 0.01 & 0.30 \\
Lift Mug & 1.57 & 4.71 & -0.40 & 0.40 & 0.44 & 0.50 & 0.10 & 0.15 \\
\hline
\end{tabular}
}
\vspace{-0.5cm}
\end{table}

\end{document}